\documentclass[runningheads, envcountsame, a4paper]{llncs}
\usepackage[pdftex]{graphicx}
\usepackage{epstopdf}
\usepackage[hyphens]{url}
\usepackage{microtype}

\usepackage[utf8]{inputenc} % allow utf-8 input
\usepackage[T1]{fontenc}    % use 8-bit T1 fonts
\usepackage{hyperref}       % hyperlinks
\usepackage{url}            % simple URL typesetting
\usepackage{booktabs}       % professional-quality tables
\usepackage{amsfonts}       % blackboard math symbols
\usepackage{nicefrac}       % compact symbols for 1/2, etc.
\usepackage{microtype}      % microtypography
\usepackage{wrapfig}
\usepackage{lipsum}
\usepackage{pifont}
\usepackage{amssymb}
\usepackage{mathrsfs}
\usepackage{subcaption}
\usepackage{times}
\usepackage{epsfig}
\usepackage{graphicx}
\usepackage{amssymb}
\usepackage[utf8]{inputenc} % allow utf-8 input
\usepackage[T1]{fontenc}    % use 8-bit T1 fonts
\usepackage{url}            % simple URL typesetting
\usepackage{booktabs}       % professional-quality tables
\usepackage{array,multirow}
\usepackage{amsfonts}       % blackboard math symbols
\usepackage{nicefrac}       % compact symbols for 1/2, etc.
\usepackage{microtype}      % microtypography
\usepackage{booktabs} % for professional tables
\usepackage{amsmath,amsfonts,graphicx,mathrsfs,cite,url}
\usepackage{tcolorbox}
\usepackage{pgfplots}
\usepackage{tikz}
\usepackage{tikz-3dplot}
\usepackage{tikzscale}
\usetikzlibrary{arrows,shapes,chains,matrix,positioning}
\usetikzlibrary{scopes,decorations,shadows,backgrounds,fit}
\usetikzlibrary{decorations.pathreplacing,calc,3d,quotes}
\usepackage{pgfmath}
\usepackage{threeparttable}
\usepackage[misc]{ifsym}
\newcommand{\M}{\mathcal M}

\newcommand{\loss}{\mathcal L}

\newcommand{\R}{\mathbb {R}}
\newcommand{\one}{{\bf{1}}}

\newcommand{\E}{\mathbb {E}}

\newcommand{\nn}{\nonumber}

\newcommand{\tr}{\mathrm{Tr}}

\usepackage{bm}
\newcommand{\J}{J}

\newcommand{\x}{{\bm x}}
\newcommand{\z}{{\bm z}}
\newcommand{\s}{{\bm s}}
\newcommand{\y}{{\bm y}}
\newcommand{\yh}{{\bm \hat y}}

\newcommand{\sh}{{\bm \hat s}}
\newcommand{\bb}{{\bm b}}
\newcommand{\bM}{{\bm M}}
\newcommand{\bK}{{\bm K}}
\newcommand{\bt}{{\bm \Theta}}

\newcommand{\bW}{{\bm W}}
\newcommand{\bI}{{\bm I}}

\newcommand{\bS}{{\bm S}}
\newcommand{\bY}{{\bm Y}}
\newcommand{\bX}{{\bm X}}
\newcommand{\bZ}{{\bm Z}}

\newcommand{\bD}{{\bm D}}

\newcommand{\bu}{{\bm u}}

\newcommand{\bB}{{\bm B}}

\newcommand{\bp}{{\bm \Phi}}
\newcommand{\bl}{{\bm \Lambda}}

\begin{document}
\toctitle{}
\title{Adversarial Representation Learning With Closed-Form Solvers}

\tocauthor{}
\author{Bashir~Sadeghi \and
Lan~Wang \and
Vishnu~Naresh~Boddeti (\Letter)}
\authorrunning{Sadeghi et al.}
\institute{Michigan State University, East Lansing, MI 48823, USA \\
\email{\{sadeghib, wanglan3, vishnu\}@msu.edu}\\
\url{http://hal.cse.msu.edu}
}

\maketitle              % typeset the header of the contribution
\begin{abstract}
Adversarial representation learning aims to learn data representations for a target task while removing unwanted sensitive information at the same time. Existing methods learn model parameters iteratively through stochastic gradient descent-ascent, which is often unstable and unreliable in practice. To overcome this challenge, we adopt closed-form solvers for the adversary and target task. We model them as kernel ridge regressors and analytically determine an upper-bound on the optimal dimensionality of representation. Our solution, dubbed OptNet-ARL, reduces to a stable one one-shot optimization problem that can be solved reliably and efficiently. OptNet-ARL can be easily generalized to the case of multiple target tasks and sensitive attributes. Numerical experiments, on both small and large scale datasets, show that, from an optimization perspective, OptNet-ARL is stable and exhibits three to five times faster convergence. Performance wise, when the target and sensitive attributes are dependent, OptNet-ARL learns representations that offer a better trade-off front between (a) utility and bias for fair classification and (b) utility and privacy by mitigating leakage of private information than existing solutions.

Code is available at \url{https://github.com/human-analysis}.

\keywords{Fair machine learning \and Adversarial representation learning  \and  Closed-form solver \and Kernel ridge regression.}
\end{abstract}

\section{Introduction}
% As such, the adversary mimics the ideal inductive biases necessary for removing unwanted information from the representation.

Adversarial Representation Learning (ARL) is a promising framework that affords explicit control over unwanted information in learned data representations. This concept has practically been employed in various applications, such as, learning unbiased and fair representations~\cite{louizos2015variational, madras2018learning,creager2019flexibly, song2018learning}, learning controllable representations that are invariant to sensitive attributes~\cite{xie2017controllable, moyer2018invariant}, mitigating leakage of sensitive information~\cite{roy2019mitigating,edwards2015censoring, sadeghi2019global,sadeghi2020imparting}, unsupervised domain adaption\cite{ganin2016domain}, learning flexibly fair representations~\cite{creager2019flexibly, song2018learning}, and many more.

\begin{figure}[t]
\centering
\begin{subfigure}{0.47\textwidth}
\centering
\includegraphics[width=\textwidth]{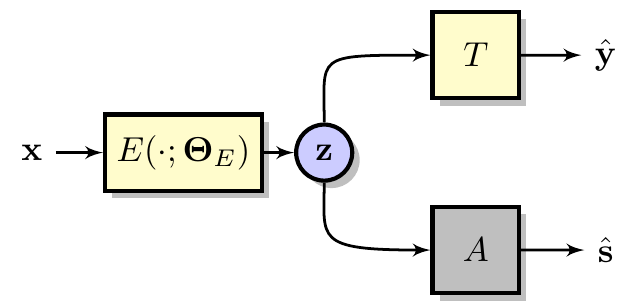}
\caption{\label{fig:overview}}
\end{subfigure}
\begin{subfigure}{0.50\textwidth}
    \centering
    \includegraphics[width=\textwidth]{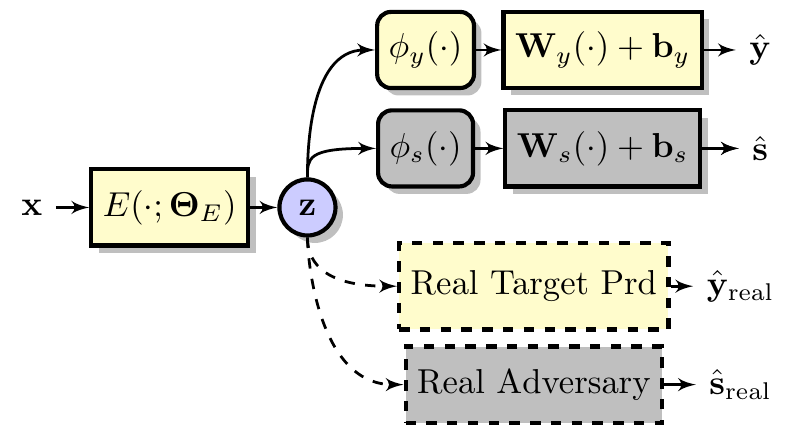}
    \caption{\label{fig:arl}}
\end{subfigure}
\caption{\textbf{Adversarial Representation Learning:} (a) Consists of three players, an encoder $E$ that obtains a compact representation $\z$ of input data $\x$, predictors $T$ and $S$ that seek to extract a desired target $\y$ and sensitive $\s$ attribute, respectively from the embedding. (b) OptNet-ARL adopts kernel regressors as proxy target predictor and adversary for learning the encoder. The learned encoder is evaluated against a real target predictor and adversary, which potentially can be neural networks.}
\end{figure}

The goal of ARL is to learn a data encoder $E:\bm{x}\mapsto\bm{z}$ that retains sufficient information about a desired \emph{target} attribute, while removing information about a known \emph{sensitive} attribute. The basic idea of ARL is to learn such a mapping under an adversarial setting. The learning problem is setup as a three-player minimax game between three entities (see Fig.\ref{fig:overview}, an encoder $E$, a predictor $T$, and a proxy adversary $A$. Target predictor $T$ seeks to extract target information and make correct predictions on the target task. The proxy adversary $A$ mimics an unknown real adversary and seeks to extract sensitive information from learned representation. As such, the proxy adversary serves only to aid the learning process and is not an end goal by itself. Encoder $E$ seeks to simultaneously aid the target predictor and limit the ability of the proxy adversary to extract sensitive information from the representation $\bm{z}$. By doing so, the encoder learns to remove sensitive information from the representation. In most ARL settings, while the encoder is a deep neural network, the target predictor and adversary are typically shallow neural networks.

The vanilla algorithm for learning the parameters of the encoder, target and adversary networks is gradient descent-ascent (GDA)~\cite{xie2017controllable, roy2019mitigating}, where the players take a gradient step simultaneously. However, applying GDA, including its stochastic version, is not an optimal strategy for ARL and is known to suffer from many drawbacks. Firstly, GDA has undesirable convergence properties; it fails to converge to a local minimax and can converge to fixed points that are not local minimax, while being very unstable and slow in practice~\cite{daskalakis2018limit,jin2019local}. Secondly, GDA exhibits strong rotation around fixed points, which requires using very small learning rates \cite{mescheder2017numerics,balduzzi2018mechanics} to converge. Numerous solutions~\cite{mescheder2017numerics,nagarajan2017gradient,gidel2018variational} have been proposed recently to address the aforementioned computational challenges. These approaches, however, seek to obtain solutions to the minimax optimization problem in the general case, where each player is modeled as a complex neural network.

In this paper, we take a different perspective and propose an alternative solution for adversarial representation learning. Our key insight is to replace the shallow neural networks with other analytically tractable models with similar capacity. We propose to adopt simple learning algorithms that admit closed-form solutions, such as linear or kernel ridge regressors for the target and adversary, while modeling the encoder as a deep neural network. Crucially, such models are particularly suitable for ARL and afford numerous advantages, including, (1) closed-form solution allows learning problems to be optimized globally and efficiently, (2) analytically obtain upper bound on optimal dimensionality of the embedding $\bm{z}$, (3) the simplicity and differentiability allows us to backpropagate through the closed-form solution, (4) practically it resolves the notorious rotational behaviour of iterative minimax gradient dynamics, resulting in a simple optimization that is empirically stable, reliable, converges faster to a local optima, and ultimately results in a more effective encoder $E$.

We demonstrate the practical effectiveness of our approach, dubbed OptNet-ARL, through numerical experiments on an illustrative toy example, fair classification on UCI Adult and German datasets and mitigating information leakage on the CelebA dataset. We consider two scenarios where the target and sensitive attributes are (a) dependent, and (b) independent. Our results indicate that, in comparison to existing ARL solutions, OptNet-ARL is more stable and converges faster while also outperforming them in terms of accuracy, especially in the latter scenario.

\noindent \textbf{Notation:} Scalars are denoted by regular lower case or Greek letters, e.g., $n$, $\lambda$. Vectors are boldface lowercase letters, e.g., $\x$, $\y$; Matrices are uppercase boldface letters, e.g., $\bX$.  A $n\times n$ identity matrix is denoted by $\bI$, sometimes with a subscript indicating its size, e.g., $\bI_n$. Centered (mean subtracted w.r.t columns) data matrix is indicated by "\textasciitilde", e.g., $\tilde{\bX}$. Assume that $\bX$ contains $n$ columns, then $\tilde{\bX} = \bX \bD$ where $\bD = \bI_n - \frac{1}{n}\one \one^T$ and $\one$ denotes a vector of ones with length of $n$. Given matrix $\bM\in \R^{m\times m}$, we use $\tr[\bM]$ to denote its trace (i.e., the sum of its diagonal elements); its Frobenius norm is denoted by $\|\bM\|_F$, which is related to the trace as $\|\bM\|_F^2 =\tr[\bM \bM^T]$. The pseudo-inverse of $\bM$ is denoted by $\bM^\dagger$. The subspace spanned by the columns of $\bM$ is denoted by $\mathcal R(M)$ or simply  $\M$ (in calligraphy); the orthogonal complement of $\M$ is denoted by $\M^\perp$. The orthogonal projector onto $\M$ is denoted by $P_\M$.

\section{Prior Work}

\noindent\textbf{Adversarial Representation Learning:} The basic idea of learning data representations with controllable semantic information has been effective across multiple topics. Domain adaptation~\cite{ganin2015unsupervised, ganin2016domain, tzeng2017adversarial}, where the goal is to learn representations that are invariant to the domain, is one of the earliest applications of ARL. More recently, adversarial learning has been extensively used \cite{edwards2015censoring,beutel2017data,xie2017controllable,zhang2018mitigating,song2018learning,elazar2018adversarial,bertran2019adversarially,roy2019mitigating,creager2019flexibly} and advocated \cite{madras2018learning} for the task of learning fair, invariant or privacy preserving representations of data. All of the aforementioned approaches represent each entity in ARL by neural networks and optimize their parameters through stochastic gradient descent-ascent (SGDA). As we show in this paper, SGDA is unstable and sub-optimal for learning. Therefore, we trade-off model expressively for ease of learning through a hybrid approach of modeling the encoder by a deep neural network and target and adversary with closed-form regressors. Such a solution reduces alternating optimization into a simple optimization problem which is much more stable, reliable and effective. Table~\ref{tbl-comparison} shows a comparative summary of ARL approaches.

\noindent\textbf{Optimization in Minmax Games:} A growing class of learning algorithms, including ARL, GANs etc., involve more than one objective and are trained via games played by cooperating or dueling neural networks. An overview of the challenges presented by such algorithms and a plausible solution in general $n$-player games can be found in \cite{letcher2019differentiable}. In the context of two-player minimax games such as GANs, a number of solutions~\cite{mescheder2017numerics,nagarajan2017gradient,balduzzi2018mechanics,daskalakis2018limit,jin2019local,gidel2018variational} have been proposed to improve the optimization dynamics, many of them relying on the idea of taking an extrapolation step \cite{korpelevich1976extragradient}. 
For example,~\cite{mescheder2017numerics} deploys some regularizations to encourage  agreement between different players and improve the convergence properties.
In another example,~\cite{nagarajan2017gradient} uses double gradient to stabilize the optimization procedure. 
In contrast to all of these approaches, that work with the given fixed models for each player, we seek to change the model of the  players in the ARL setup for ease of optimization. In the context of ARL, \cite{sadeghi2019global} considers the setting where all the players, including the encoder, are linear regressors. While they obtained a globally optimum solution, the limited model capacity hinders the flexibility (cannot directly use raw data), scalability and performance (limited by pre-trained features) of their solution. In this paper we advocate the use of ridge regressors (linear or kernel) for the target and adversary, while modeling the encoder as a deep neural network. This leads to a problem that obviates the need for gradient descent-ascent and can instead be easily optimized with standard SGD. Not only does this approach lead to stable optimization, it also scales to larger datasets and exhibits better empirical performance.

\begin{table}[t]
\centering
\caption{Comparison between different ARL methods ($n$: sample size, $b$: batch size).}
\resizebox{\textwidth}{!}{
\begin{tabular}{ p{3.0cm} p{4.9cm} p{2.25cm} p{1.4cm} p{1.95cm} p{1.7cm}}
\toprule
\textbf{\hfil Method} &\textbf{Encoder / Target \& Adversary}& \hfil \textbf{Optimization}& \textbf{Scalability}   & \textbf{\hfil Enc Soln}& \textbf{Input Data}\\
\midrule
 \hfil \textbf{SGDA-ARL}~\cite{xie2017controllable,madras2018learning} & \hfil deep NN / shallow NN &  alternating SGD &  \hfil $\geq \mathcal{O}(b^3)$& \hfil unknown & \hfil raw data\\ 
 \textbf{Kernel-SARL}~\cite{sadeghi2019global} & \hfil kernel regressor / linear & \hfil closed-form&  \hfil $\mathcal{O}(n^3)$ & \hfil global optima & \hfil features\\  
 \midrule
 \textbf{\hfil OptNet-ARL} (ours) &  \hfil deep NN / kernel regressor  & \hfil SGD & \hfil $\mathcal{O}(b^3)$ & \hfil local optima & \hfil raw data\\
 \toprule
\end{tabular}\label{tbl-comparison}
}
\end{table}

\noindent\textbf{Differentiable Solvers:} A number of recent approaches have integrated differentiable solvers, both iterative as well as closed-form, within  end-to-end learning systems. Structured layers for segmentation and higher order pooling were introduced by \cite{ionescu2015training}. Similarly \cite{valmadre2017end} proposed an asymmetric architecture which incorporates a Correlation Filter as a differentiable layer. Differential optimization as a layer in neural networks was introduced by \cite{amos2017optnet,agrawal2019differentiable}. More recently, differentiable solvers have also been adopted for meta-learning \cite{bertinetto2018meta,lee2019meta} as well. The primary motivation for all the aforementioned approaches is to endow deep neural networks with differential optimization and ultimately achieve faster convergence of the end-to-end system. In contrast, our inspiration for using differential closed-form solvers is to control the non-convexity of the optimization in ARL, in terms of stability, reliability and effectiveness.

\section{Problem Setting}

Let the data matrix $\bX=[\x_1,\dots, \x_n] \in {\R}^{d\times n}$ be $n$ realizations of $d$-dimensional data, $\x\in \R^d$.
Similarly, we denote $n$ realizations of sensitive attribute vector $\s\in \R^q$ and target attribute vector $\y\in\R^p$ by matrices $\bS=[\s_1,\cdots,\s_n]$ and $\bY=[\y_1,\cdots,\y_n]$, respectively. Treating the attributes as vectors enables us to consider both multi-class classification and regression under the same formulation. Each data sample $\x_k$ is associated with the sensitive attribute $\s_k$  and the target attribute $\y_k$, respectively.

The ARL problem is formulated with the goal of learning parameters of an embedding function $E(\cdot;\bt_E)$ that maps a data sample $\x$ to $\z\in\R^r$ with two objectives: (i) aiding a target predictor $T(\cdot;\bt_y)$ to accurately infer the target attribute $\y$ from $\z$, and (ii) preventing an adversary $A(\cdot;\bt_s)$ from inferring the sensitive attribute $\s$ from $\z$. The ARL problem can be formulated as a bi-level optimization, 
\begin{equation}
    \begin{aligned}
        \min_{\bt_E} \min_{\bt_y}  \mbox{ } \mathcal{L}_y\left( T(E(\x;\bt_E);\bt_y), \y \right) \quad
        \mathrm {s.t. \ \  }  \min_{\bt_s} \mathcal{L}_s\left( A(E(\x;\bt_E);\bt_s), \s \right) \geq \alpha
    \end{aligned}
    \label{eq:arl}
\end{equation}
\noindent where $\mathcal{L}_y$ and $\mathcal{L}_s$ are the loss functions (averaged over the training dataset) for the target predictor and the adversary, respectively; $\alpha \in (0,\infty)$ is a user defined value that determines the minimum tolerable loss $\alpha$ for the adversary on the sensitive attribute; and the minimization in the constraint is equivalent to the encoder operating against an optimal adversary. Denote the global minimums of the adversary and target estimators as
\begin{equation}
    \begin{aligned}
    \label{eq:min_MSEs}
    J_y (\bt_E) &:= \min_{\bt_y}\loss_y \left( T(E(\x;\bt_E);\bt_y), \y \right)\\
    J_s (\bt_E) &:= \min_{\bt_s} \loss_s\left( A(E(\x;\bt_E);\bt_s), \s \right).
    \end{aligned}
\end{equation}
The constrained optimization problem in~(\ref{eq:arl}) can be alternately solved through its Lagrangian version:
\begin{equation}
\min_{\bt_E}  \Big\{(1-\lambda) J_y (\bt_E) -\lambda J_s (\bt_E) \Big\} , \ 0\le\lambda\le 1.
\label{eq:main}
\end{equation}

\subsection{Motivating Exact Solvers}
Most state-of-the-art ARL algorithms cannot solve
the optimization problems in~(\ref{eq:min_MSEs}) optimally (e.g., SGDA).
For any given $\bt_E$, denote the non-optimal adversary and target predictors loss
functions as $J_y^{\text{approx}}(\bt_E)$ and $J_s^{\text{approx}}(\bt_E)$, respectively.
It is obvious that for any given $\bt_E$, it holds
\begin{eqnarray}
J_y^{\text{approx}}(\bt_E) \ge J_y(\bt_E) \ \ \ \text{and} \ \ \
J_s^{\text{approx}}(\bt_E) \ge J_s(\bt_E).\nn
\end{eqnarray}
Note that the optimization problem raised from non-optimal adversary and target predictors is
\begin{equation}
\min_{\bt_E}  \Big\{(1-\lambda) J_y^{\text{approx}} (\bt_E) -\lambda J_s^{\text{approx}} (\bt_E) \Big\} \ , 0\le\lambda\le 1.
\label{eq:main-sub}
\end{equation}

Intuitively, solution(s) of~(\ref{eq:main-sub}) do not outperform that of~(\ref{eq:main}). We now formulate this intuition more concretely. 
\begin{definition}
Let $(a_1, a_2)$ and $(b_1, b_2)$ be two arbitrary points in $\R^2$. We say $(b_1, b_2)$ dominates $(a_1, a_2)$ if and only if $b_1> a_1$ and $b_2<a_2$ hold simultaneously.
\end{definition}

\begin{theorem}

For any $\lambda_1,\, \lambda_2 \in [0,1]$, consider the following optimization problems
\begin{equation}\label{eq:theta-exact}
\bt_E^\text{exact} =\arg\min_{\bt_E} \Big\{(1-\lambda_1) J_y (\bt_E) -\lambda_1 J_s (\bt_E) \Big\}
\end{equation}
and
\begin{equation}
\bt_E^\text{approx} =\arg\min_{\bt_E} \Big\{(1-\lambda_2) J_y^{\text{approx}} (\bt_E) -\lambda_2 J_s^{\text{approx}} (\bt_E) \Big\}\nn
\end{equation}
Then, any adversary-target objective trade-off generated by $\big(J_s(\bt_E^\text{exact}), J_y(\bt_E^\text{exact}) \big)$ cannot be dominated by the trade-off generated by $\big(J_s(\bt_E^\text{approx}), J_y(\bt_E^\text{approx})\big)$.
\end{theorem}
See supplementary material for the proof of all Lemmas and Theorems.

\section{Approach}
Existing instances of ARL adopt deep neural networks to represent $E$, $T$ and $A$ and learn their respective parameters $\{\bt_E, \bt_y, \bt_s\}$ through stochastic gradient descent-ascent (SGDA). Consequently, the target and adversary in Eq. \ref{eq:min_MSEs} are not solved to optimality, thereby resulting in a sub-optimal encoder.

\subsection{Closed-Form Adversary and Target Predictor\label{sec:kernelized-regrossor}}
The machine learning literature offers a wealth of methods with exact solutions that are appropriate for modeling both the adversary and target predictors. In this paper, we argue for and adopt simple, fast and differentiable methods such as kernel ridge regressors as shown in Fig. \ref{fig:arl}. On one hand, such modeling allows us to obtain the optimal estimators globally for any given encoder $E(\cdot;\bt_E)$.

On the other hand, kernelized ridge regressors can be stronger than the shallow neural networks that are used in many ARL-based solutions (e.g.,~\cite{xie2017controllable,elazar2018adversarial,madras2018learning,roy2019mitigating}). Although it is not the focus of this paper, it is worth noting that even deep neural networks in the infinite-width limit reduce to linear models with a kernel called the neural tangent kernel~\cite{jacot2018neural}, and as such can be adopted to increase the capacity of our regressors.

Consider two reproducing kernel Hilbert spaces (RKHS) of functions $\mathcal{H}_s$ and $\mathcal{H}_y$ for adversary and target regressors, respectively. Let a possible corresponding pair of feature maps be $\phi_s(\cdot)\in\R^{r_s}$ and $\phi_y(\cdot)\in\R^{r_y}$ where $r_s$ and $r_y$ are the dimensionality of the resulting features and can potentially approach infinity. The respective kernels for $\mathcal{H}_s$ and $\mathcal{H}_y$ can be represented as $k_s(\z_1, \z_2) = \langle \phi_s(\z_1), \phi_s(\z_2) \rangle_{\mathcal{H}_s}$ and $k_y(\z_1, \z_2) = \langle \phi_y(\z_1), \phi_y(\z_2) \rangle_{\mathcal{H}_y}$. Under this setting, we can relate the target and sensitive attributes to any given embedding $\z$ as,
\begin{eqnarray}\label{estimators-kernel}
\bm{\yh} = \bW_y \phi_y(\z) + \bb_y, \qquad  \bm{\sh} = \bW_s \phi_s(\z) + \bb_s
\end{eqnarray}
where $\bt_y=\{\bW_y,\bb_y\}$ and $\bt_s=\{\bW_s,\bb_s\}$ are the regression parameters, $\bW_y \in \R^{p\times r_y}$ and $\bW_s \in \R^{q\times r_s}$, $\bb_y \in \R^p$ and $\bb_s \in \R^q$ respectively.

Let the entire embedding of input data be denoted as $\bZ:= [\z_1,\cdots,\z_n]$ and the corresponding features maps as  $\bp_y:= [\phi_y(\z_1),\cdots,\phi_y(\z_n)]$ and $\bp_s:= [\phi_s(\z_1),\cdots,\phi_s(\z_n)]$, respectively.
Furthermore, we denote the associated Gram matrices by ${\bK}_y  = {\bp_y} ^T {\bp_y}$ and ${\bK}_s= {\bp_s} ^T {\bp_s}$. A centered Gram matrix $\tilde{\bK}$ corresponding to the Gram matrix $\bK$ can be obtained \cite{gretton2005kernel} as,
 \begin{equation}
     \tilde{\bK} = \tilde{\bp} ^T \tilde{\bp} = (\bp \bD)^T(\bp \bD) = \bD^T\bK \bD.
 \end{equation}
Invoking the representer theorem~\cite{shawe2004kernel}, the regression parameters can be represented as $\bW_y = \bl_y \tilde{\bp}_y^T$ and $\bW_s = \bl_s \tilde{\bp}_s^T$ for target and adversary respectively, where $\bl_y\in\R^{p\times n}$ and $\bl_s\in\R^{n\times q}$ are new parameter matrices. As a result, the kernelized regressors in~(\ref{estimators-kernel}) can be equivalently expressed as
\begin{eqnarray}\label{estimators-kernel-representor}
\yh = \bl_y \tilde{\bp}_y^T\phi_y(\z)+\bb_y, \quad  \sh = \bl_s \tilde{\bp}_s^T\phi_s(\z)+ \bb_s.
\end{eqnarray}

In a typical ARL setting, once an encoder is learned (i.e., for a given fixed embedding $\z$), we evaluate against the best possible adversary and target predictors.
% Therefore, we first consider the embedding in isolation from the encoder and analyze the downstream ridge regressors.
In the following Lemma, we obtain the minimum MSE for kernelized adversary and target predictors for any given embedding $\bZ$.
\begin{lemma}
Let $J_y(\bZ)$ and $J_s(\bZ)$ be regularized minimum MSEs for adversary and target:
\begin{eqnarray}\label{eq-J-kernel}
&&J_y(\bZ) = \min_{\bl_y, \bb_y} \Big\{\E \big\{ \big\|\hat \y - \y\big\|^2\big\} + \gamma_y  \big\| \bl_y\big\|_F^2\Big\},\nn\\
&& J_s(\bZ) = \min_{\bl_s, \bb_s} \Big\{\E \big\{ \big\|\hat\s - \s\big\|^2\big\} + \gamma_s  \big\| \bl_s\big\|_F^2\Big\}\nn
\end{eqnarray}
where $\gamma_y$ and $\gamma_s$ are regularization parameters for target and adversary regressors, respectively.
Then, for any given embedding matrix $\bZ$, the minimum MSE for kernelized adversary and target can be obtained as
\begin{eqnarray}\label{eq:J-kernel}
&&\J_y (\bZ) =  \frac{1}{n}\big\|\tilde{\bY} \big\|_F^2-\frac{1}{n}\bigg\| P_{\M_y}
\begin{bmatrix}
\tilde{\bY}^T  \\
\mathbf{0}_{n}
\end{bmatrix}
\bigg\|_F^2,\nn\\
&& \J_s (\bZ) =  \frac{1}{n}\big\|\tilde{\bS} \big\|_F^2-\frac{1}{n}\bigg\| P_{\M_s}
\begin{bmatrix}
\tilde{\bS}^T  \\
\mathbf{0}_{n}
\end{bmatrix}
\bigg\|_F^2
\end{eqnarray}
where
\begin{equation}
\bM_y = \begin{bmatrix}
\tilde{\bK}_y  \\
\sqrt{n\gamma_y}\bI_{n}
\end{bmatrix}
, \qquad \ \bM_s = \begin{bmatrix}
\tilde{\bK}_s  \\
\sqrt{n\gamma_s}\bI_{n}\nn
\end{bmatrix}
\end{equation}
are both full column rank matrices and
 a projection matrix for any full column rank matrix $\bM$ is
\begin{equation}
    P_\M = \bM (\bM^T \bM)^{-1} \bM^T\nn
\end{equation}
\end{lemma}

It is straightforward to generalize this method to the case of multiple
target and adversary predictors through equation (3). In this case we will
have multiple $\lambda$'s to trade-off between fairness and utility. 

\subsection{Optimal Embedding Dimensionality}
% To do this, we disconnect the embedding vector $\z$ from the encoder in Figure~\ref{fig:arl} and
% let $\z$ be a free vector in $\R^r$. Our approximation objective is to aid maximally our target predictor as
% long as the embedding dimensionality is concerned. Therefore, we model the target predictor as a simple linear regressor.
% In order to find an upper bound for the considered approximation, we also model adversary vi a linear regressor since
% any stronger adversary requires smaller embedding dimensionality. Thus, the following theorem presents the approximate embedding dimensionality.
% In Section 6, we show that, practically, the embedding dimensionality plays a central role in hindering the leakage of sensitive attributes from the representation in both existing ARL approaches as well as our OptNet-ARL solution.

The ability to effectively optimize the parameters of the encoder is critically dependent on the dimensionality of the embedding as well. Higher dimensional embeddings can inherently absorb unnecessary extraneous information in the data. Existing ARL applications, where the target and adversary are non-linear neural networks, select the dimensionality of the embedding on an ad-hoc basis.

Adopting closed-form solvers for the target and adversary enables us to analytically determine an upper bound on the optimal dimensionality of the embedding for OptNet-ARL. To obtain the upper bound we rely on the observation that a non-linear target predictor and adversary, by virtue of greater capacity, can learn non-linear decision boundaries. As such, in the context of ARL, the optimal dimensionality required by non-linear models is lower than the optimal dimensionality of linear target predictor and adversary. Therefore, we analytically determine the optimal dimensionality of the embedding in the following theorem.

\begin{theorem}
Let $\bm z$ in Figure~\ref{fig:arl} be disconnected from the encoder and be a free vector in $\mathbb R^r$. Further, assume that both adversary and target predictors are linear regressors. Then, for any $0 \le \lambda \le 1$ the optimal dimensionality of embedding vector, $r$ is the number of negative eigenvalues of 
\begin{eqnarray}\label{B}
\bB &=& \lambda \tilde{\bS}^T \tilde{\bS} -(1-\lambda)\tilde{\bY}^T \tilde{\bY}.
\end{eqnarray}
\label{thm2}
\end{theorem}

Given a dataset with the target and sensitive labels, $\bm{Y}$ and $\bm{S}$ respectively, the matrix $\bm{B}$ and its eigenvalues can be computed offline to determine the upper bound on the optimal dimensionality.
By virtue of the greater capacity, the optimal dimensionality required by non-linear models is lower than the optimal dimensionality of linear
predictors and therefore, Theorem $2$ is a tight upper bound
for the optimal embedding dimensionality.
One large datasets where $\bm{B}\in\mathbb{R}^{n\times n}$, the Nystr\"{o}m method with data sampling \cite{kumar2012sampling} can be adopted.

% At the first glance this approximation may not seem very practical for non-linear target predictor and adversary. However, this approximation represents an upper bound for the optimal embedding dimensionality since any stronger target predictor and adversary ( compared to linear regressor), require smaller embedding dimensionality.

\subsection{Gradient of Closed-Form Solution}
In order to find the gradient of the encoder loss function in~(\ref{eq:main}) with $J_y$ and $J_s$ given in~(\ref{eq:J-kernel}), we can ignore the constant terms, $\|\tilde{\bY}\|_F$ and $\|\tilde{\bS}\|_F$. Then, the optimization problem in~(\ref{eq:main}) would be equivalent to
\begin{align}\label{eq:implemneted}
\min_{\bt_E} \Bigg\{(1-\lambda)\bigg\| P_{\M_s}
\begin{bmatrix}
\tilde{\bS}^T  \\
\mathbf{0}_{n}
\end{bmatrix}
\bigg\|_F^2
-\lambda \bigg\| P_{\M_y}
\begin{bmatrix}
\tilde{\bY}^T \nn \\
\mathbf{0}_{n}
\end{bmatrix}
\bigg\|_F^2\Bigg\} \\
=\min_{\bt_E}\Bigg\{(1-\lambda)\sum_{k=1}^p\|P_{\M_s}\bu_s^k\|^2-\lambda \sum_{m=1}^q\|P_{\M_y}\bu_y^m\|^2\Bigg\}
\end{align}
where the vectors $\bu_s^k$ and $\bu_y^m$ are the $k$-th and $m$-th columns of  
$\begin{bmatrix}
\tilde{\bS}^T  \\
\mathbf{0}_{n}
\end{bmatrix}$ 
and 
$\begin{bmatrix}
\tilde{\bY}^T  \\
\mathbf{0}_{n}
\end{bmatrix}$, respectively. 
Let $\bM$ be an arbitrary matrix function of $\bt_E$, and $\theta$ be arbitrary scalar element of $\bt_E$. Then, from~\cite{golub1973differentiation} we have
\begin{eqnarray}\label{eq:derivative}
\frac{1}{2} \frac{\partial \|P_{\M} \bu\|^2}{\partial \theta}  =  \bu^T P_{\M^\perp} \frac{\partial \bM }{\partial \theta} \,\bM^{\dagger} \bu
\end{eqnarray}
where
{\footnotesize
\begin{eqnarray}
\Big[\frac{\partial \bM }{\partial \theta}\Big]_{ij}\nn = \begin{cases} \nabla^T_{\z_i}\big([\bM]_{ij}\big)
\nabla_\theta({\z_i})
+\nabla^T_{\z_j}\big([\bM]_{ij}\big)
\nabla_\theta({\z_j}), & i\le n\\0, & \text{else}.
\end{cases}
\end{eqnarray}}
Equation~(\ref{eq:derivative}) can be directly used to obtain the gradient of objective function in~(\ref{eq:implemneted}).

Directly computing the gradient in Eq~(\ref{eq:derivative}) requires a pseudoinverse of the matrix $\bm{M}\in\mathbb{R}^{2n\times n}$, which has a complexity of $\mathcal{O}(n^3)$. For large datasets this computation can get prohibitively expensive. Therefore, we approximate the gradient using a single batch of data as we optimize the encoder end-to-end. Similar approximations~\cite{kumar2012sampling} are in fact commonly employed to scale up kernel methods. Thus, the computational complexity of computing the loss for OptNet-ARL reduces to $\mathcal{O}$($b^3$), where $b$ is the batch size. Since maximum batch sizes in training neural networks are of the order of 10s to 1000s, computing the gradient is practically feasible.
We note that, the procedure presented in this section is a simple SGD in which its stability can be guaranteed under Lipschitz and smoothness assumptions on encoder network~\cite{hardt2016train}.

\section{Experiments}
\label{sec-experiment}
In this section we will evaluate the efficacy of our proposed approach, OptNet-ARL, on three different tasks; Fair Classification on UCI~\cite{dua2017uci} datatset, mitigating leakage of private information on the CelebA dataset, and ablation study on a Gaussian mixture example. We also compare OptNet-ARL with other ARL baselines in terms of stability of optimization, the achievable trade-off front between the target and adversary objectives, convergence speed and the effect of embedding dimensionality. We consider three baselines, (1) \textbf{SGDA-ARL:} vanilla stochastic gradient descent-ascent that is employed by multiple ARL approaches including \cite{xie2017controllable,elazar2018adversarial,madras2018learning,roy2019mitigating,kim2019learning} etc., (2) \textbf{ExtraSGDA-ARL:} a state-of-the-art variant of stochastic gradient descent-ascent that uses an extra gradient step \cite{korpelevich1976extragradient} for optimizing minimax games. Specifically, we use the ExtraAdam algorithm from \cite{gidel2018variational}, and (3) \textbf{SARL:} a global optimum solution for a kernelized regressor encoder and linear target and adversary~\cite{sadeghi2019global}. 
Specifically, \textbf{hypervolume} (HV) \cite{zitzler1998multiobjective},  a metric for stability and goodness of trade-off (comparing algorithms under multiple objectives) is also utilized.
A larger HV indicates a better Pareto front achieved and the standard deviation of the HV represents the stability.

% Each experiment that we conduct consists of two stages. 
In the training stage, the encoder, a deep neural network, is optimized \textbf{end-to-end} against kernel ridge regressors (RBF Gaussian kernel\footnote{$k(\bm z, \bm z^\prime)=\exp{(-\frac{\|\bm z-\bm z^\prime\|^2)}{2\sigma^2}}$}) in the case of OptNet-ARL and multi-layer perceptrons (MLPs) for the baselines. Table~\ref{tbl-settings} summarizes the network architecture of all experiments.
We note that the optimal embedding dimensionality, $r$ for binary target is equal to one which is consistent with Fisher's linear discriminant analysis~\cite{fisher1936use}.
The embedding is instance normalized (unit norm). So we adopted a fixed value of $\sigma=1$ for Gaussian Kernel in all the experiments. We let the regression regularization parameter be $10^{-4}$ for all experiments. The learning rate is $3\times10^{-4}$ with weight decay of $2\times10^{-4}$ and we use Adam as optimizer for all experiments. 

At the inference stage, the encoder is frozen, features are extracted and a new target predictor and adversary are trained. At this stage, for both OptNet-ARL and the baselines, the target and adversary have the same model capacity. Furthermore, each experiment on each dataset is repeated five times with different random seeds (except for SARL which has a closed-form solution for encoder) and for different trade-off parameters $\lambda\in[0,1]$. We report the median and standard deviation across the five repetitions.
\begin{table}[t]
\caption{Network Architectures in Experiments. }
    \centering
\scalebox{0.8}{
\begin{tabular}{p{3.3cm}| c c c c c c}
\toprule

     {Method} & Encoder & Embd & Target & Adversary& Target & Adversary \\
          (ARL) &  &Dim & (Train) & (Train) & (Test) & (Test) \\
        %  & (lys), (hd) & & (lys), (hd) & (lys), (hd) & (lys), (hd) & (lys), (hd) \\
           \hline
            \hline
               & \multicolumn{6}{c}{\textbf{Adult}} \\
        %   \hline
          SGDA~\cite{xie2017controllable,madras2018learning} & MLP-$4$-$2$& $1$& MLP-$4$ & MLP-$4$& MLP-$4$-$2$& MLP-$4$-$2$\\
        ExtraSGDA~\cite{korpelevich1976extragradient}& MLP-$4$-$2$& $1$& MLP-$4$ & MLP-$4$& MLP-$4$-$2$& MLP-$4$-$2$\\
        SARL~\cite{sadeghi2019global}&RBF krnl& $1$& linear & linear& MLP-$4$-$2$& MLP-$4$-$2$\\
        OptNet-ARL (ours)& MLP-$4$-$2$& $1$& RBF krnl& RBF krnl& MLP-$4$-$2$& MLP-$4$-$2$\\
        \hline
                       & \multicolumn{6}{c}{\textbf{German}} \\
        %   \hline
          SGDA~\cite{xie2017controllable,madras2018learning} & MLP-$4$& $1$& MLP-$2$ & MLP-$2$& logistic& logistic\\
        ExtraSGDA~\cite{korpelevich1976extragradient}& MLP-$4$& $1$& MLP-$2$ & MLP-$2$&logistic&logistic\\
        SARL~\cite{sadeghi2019global}&RBF krnl& $1$& linear & linear& logistic& logistic\\
        OptNet-ARL (ours)& MLP-$4$& $1$& RBF krnl& RBF krnl& logistic& logistic\\
        \hline
               & \multicolumn{6}{c}{\textbf{CelebA}} \\
        %   \hline
          SGDA~\cite{xie2017controllable,madras2018learning} & ResNet-$18$& $128$& MLP-$64$ & MLP-$64$& MLP-$32$-$16$& MLP-$32$-$16$\\
        ExtraSGDA~\cite{korpelevich1976extragradient}& ResNet-$18$& $128$& MLP-$64$-$32$& MLP-$64$-$32$ & MLP-$32$-$16$& MLP-$32$-$16$\\
        OptNet-ARL (ours)& ResNet-$18$& $[1, 128]$& RBF krnl & RBF krnl& MLP-$32$-$16$& MLP-$32$-$16$\\
        \hline        
         & \multicolumn{6}{c}{\textbf{Gaussian Mixture}} \\
        %   \hline
          SGDA~\cite{xie2017controllable,madras2018learning} & MLP-$8$-$4$& $2$& MLP-$8$-$4$ & MLP-$8$-$4$& MLP-$4$-$4$& MLP-$4$-$4$\\
        ExtraSGDA~\cite{korpelevich1976extragradient}& MLP-$8$-$4$& $2$& MLP-$8$-$4$ & MLP-$8$-$4$& MLP-$4$-$4$& MLP-$4$-$4$\\
        SARL~\cite{sadeghi2019global}&RBF krnl& $2$& linear & linear& MLP-$4$-$4$& MLP-$4$-$4$\\
        RBF-OptNet-ARL (ours)& MLP-$8$-$4$& $2$& RBF krnl& RBF krnl& MLP-$4$-$4$& MLP-$4$-$4$\\
        IMQ-OptNet-ARL (ours)& MLP-$8$-$4$& $[1, \cdots, 512]$& IMQ krnl& IMQ krnl& MLP-$4$-$4$& MLP-$4$-$4$\\
        \hline
          \toprule
\end{tabular}\label{tbl-settings}
}
\end{table}

\subsection{Fair Classification}

We consider fair classification on two different tasks. {\bf{UCI Adult Dataset:}} It includes $14$ features from $45,222$ instances. The  task is to classify the annual income of each person as high ($50$K or above) or low (below $50$K). The sensitive feature we wish to be fair with respect to is the gender of each person. {\bf{UCI German Dataset:}} It contains $1000$ instances of individuals with $20$ different attributes. The target task is to predict their creditworthiness while being unbiased with respect to age. The correlation between target and sensitive attributes are $0.03$ and $0.02$ for the Adult and German dataset, respectively. This indicates that the target attributes are almost orthogonal to the sensitive attributes. Therefore, the sensitive information can be totally removed with only a negligible loss in accuracy for the target task.

\begin{table}[t]
    \centering
    \caption{Fair Classification On UCI Dataset (in \%)}
    \scalebox{0.8}{
    \begin{tabular}{l|c c c |c c c}
    \toprule
    &\multicolumn{3}{c}{Adult Dataset} & \multicolumn{3}{|c}{German Dataset}\\
    \midrule
    Method & Target & Sensitive & Diff& Target & Sensitive & Diff\\
    & (income)& (gender)&$67.83$&(credit)&(age)&$81$\\
    \midrule 
    Raw Data & $85.0$& $85.0$ & $17.6$& $80.0$ & $87.0$ & $6.0$\\
    \midrule
    LFR~\cite{zemel2013learning} & $82.3$ & $67.0$ & $0.4$& $72.3$& $80.5$& $0.5$\\
    AEVB~\cite{kingma2013auto} & $81.9$ & $66.0$ & $1.4$& $72.5$ & $79.5$& $1.5$\\
    VFAE~\cite{louizos2015variational} & $81.3$& $67.0$ & $0.4$ & $72.7$ & $79.7$ & $1.3$\\
    SARL~\cite{sadeghi2019global} & $84.1$& $67.4$ & $0.0$& $76.3$ &$80.9$ & $0.1$\\
    \midrule
    SGDA-ARL~\cite{xie2017controllable}& $83.61\pm0.38$ & $67.08\pm0.48$& $0.40$& $76.53\pm1.07$& $87.13\pm5.70$&$6.13$\\
    ExtraSGDA-ARL~\cite{gidel2018variational}& $83.66\pm0.26$ & $66.98\pm0.49$& $0.4$& $75.60\pm1.68$& $86.80\pm4.05$&$5.80$\\
    OptNet-ARL~& $83.81\pm0.23$ & $67.38\pm0.00$& $0.00$& $76.67\pm2.21$& $80.13\pm1.48$&$0.87$\\
    \toprule
    \end{tabular}}
    \label{tab:uci}
\end{table}

\noindent{\bf{Stability}:} Since there is no trade-off between the two attributes, we compare stability by reporting the median and standard deviation of the target and adversary performance in Table \ref{tab:uci}. Our results indicate that OptNet-ARL achieves a higher accuracy for target task and lower leakage of sensitive attribute and with less variance. For instance, in Adult dataset, OptNet-ARL method achieves $83.86\%$ and $83.81\%$ target accuracy with almost zero sensitive leakage. For OptNet-ARL the standard deviation of sensitive attribute is exactly zero, which demonstrates its effectiveness and stability in comparison to the baselines. Similarly for the German dataset, OptNet-ARL achieves $80.13\%$ for sensitive accuracy, which is close to random chance (around $81\%$).

\noindent{\bf{Fair Classification Performance}:} We compare our proposed approach with many baseline results on these datasets. The optimal dimensionality for OptNet-ARL is $r=1$ as determined by Theorem \ref{thm2} and $r=50$ for the baselines (common choice in previous work). Diff value in Table \ref{tab:uci} shows the difference between adversary accuracy and random guessing. On both datasets, both Linear-SARL and OptNet-ARL can achieve high performance on target task with a tiny sensitive attribute leakage for the German dataset.

\subsection{Mitigating Sensitive Information Leakage}
The CelebA dataset~\cite{liu2015deep} contains $202,599$ face images of $10,177$ celebrities. Each image contains 40 different binary attributes (e.g., gender, emotion, age, etc.). Images are pre-processed and aligned to a fixed size of $112 \times 96$ and we use the official train-test splits. The target task is defined as predicting the presence or absence of high cheekbones (binary) with the sensitive attribute being smiling/not smiling (binary). The choice of this attribute pair is motivated by the presence of a trade-off between them. We observe that the correlation between this attribute pair is equal to $0.45$, indicating that there is no encoder that can maintain target performance without leaking the sensitive attribute. 
 
For this experiment, we note that SARL~\cite{sadeghi2019global} cannot be employed, since, (1) it does not scale to large datasets ($\mathcal{O}(n^3)$) like CelebA, and (2) it cannot be applied directly on raw images but needs features extracted from a pre-trained network. Most other attribute pairs in this dataset either suffer from severe class imbalance or small correlation, indicating the lack of a trade-off. Network architecture details are shown in Table~\ref{tbl-settings}.

\begin{figure}[t]
\centering
\begin{subfigure}{0.44\textwidth}
\centering
\includegraphics[ width=\textwidth]{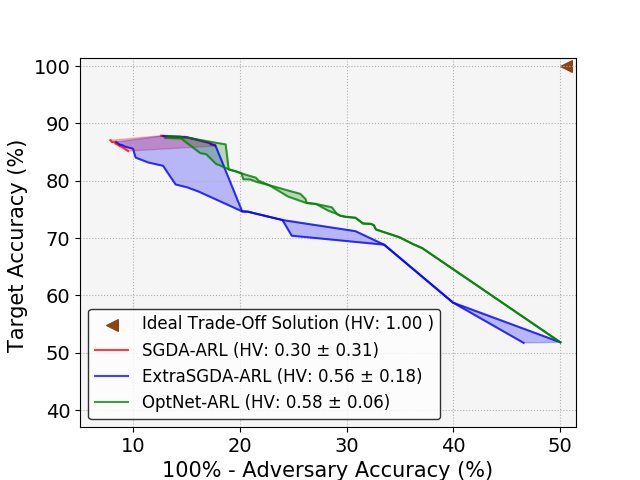}
\caption{}
\end{subfigure}
\begin{subfigure}{0.44\textwidth}
\centering
\includegraphics[ width=\textwidth]{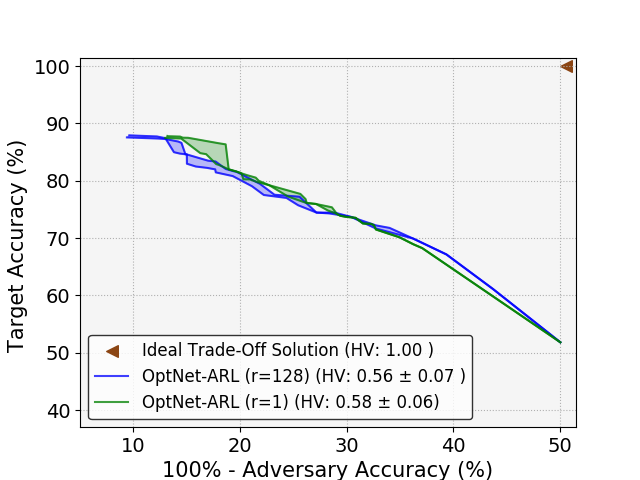}
\caption{}
\end{subfigure}
\begin{subfigure}{0.44\textwidth}
\centering
\includegraphics[width=\textwidth]{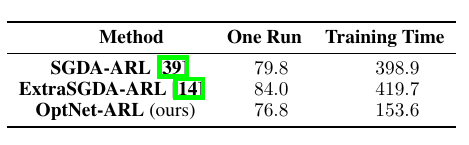}
\caption{}
\end{subfigure}
\begin{subfigure}{0.44\textwidth}
\centering
\includegraphics[ width=\textwidth]{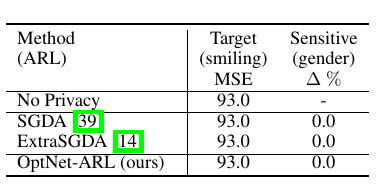}
\caption{}
\end{subfigure}
\caption{\textbf{CelebA:} (a) Trade-off between adversary and target accuracy for dependent pair (smiling/not-smiling, high cheekbones). 
(b) Comparison between the trade-offs of optimal embedding dimensionality $r=1$ and that of $r=128$.
(c) Overall and single run training time for different ARL methods.
(d) Trade-off between adversary and target for independent pair (smiling/not-smiling, gender).
}\label{fig-results1}
\end{figure}

\noindent{\textbf{Stability and Trade-off}:} Figure~\ref{fig-results1}(a) shows the attainment surface~\cite{knowles2005summary} and  hypervolume~\cite{zitzler1998multiobjective} (median and standard deviation) for all methods. SGDA spans only a small part of the trade-off and at the same time exhibits large variance around the median curves. Overall both baselines are unstable and unreliable when the two attributes are dependent on each other. On the other hand, OptNet-ARL solutions are very stable and while also achieving a better trade-off between target and adversary accuracy.

\noindent{\textbf{Optimal Embedding Dimensionality:}}
Figure~\ref{fig-results1}(b) compares the utility-bias trade-off the sub-optimal embedding dimensionality ($r=128$) with that of the optimal dimensionality ($r=1$). We can observe that optimal embedding dimensionality ($r=1$)
is producing a more stable trade-off between adversary and target accuracies.

\noindent{\textbf{Training Time:}}
It takes five runs for SGDA-ARL and ExtraSGDA and two runs for OptNet-ARL to train a reliable encoder for overall $11$ different values of $\lambda\in[0,1]$.
The summary of training time is given in Figure~\ref{fig-results1}(c). ExtraSGDA-ARL takes an extra step to update the weights and therefore, it is slightly slower than SGDA-ARL. OptNet-ARL on the other hand is significantly faster to obtain reliable results. Even for a single run, OptNet-ARL is faster than the baselines. This is because, OptNet-ARL uses closed-form solvers for adversary and target and therefore does not need to train any additional networks downstream to the encoder.

\noindent\textbf{Independent Features:} We consider the target task to be binary classification of smiling/not smiling with the sensitive attribute being gender. In this case, the correlation between gender and target feature is $0.02$, indicating that the two attributes are almost independent and hence it should be feasible for an encoder to remove the sensitive information without affecting target task. The results are presented in Figure~\ref{fig-results1} (d). In contrast to the scenario where the two attributes are dependent, we observe that all ARL methods can perfectly hide the sensitive information (gender) from representation without loss of target task. Therefore, OptNet-ARL is especially effective in a more practical setting where the target and sensitive attributes are correlated and hence can only attain a trade-off.

\subsection{Ablation Study on Mixture of Four Gaussians}
\label{sec-gaussian}
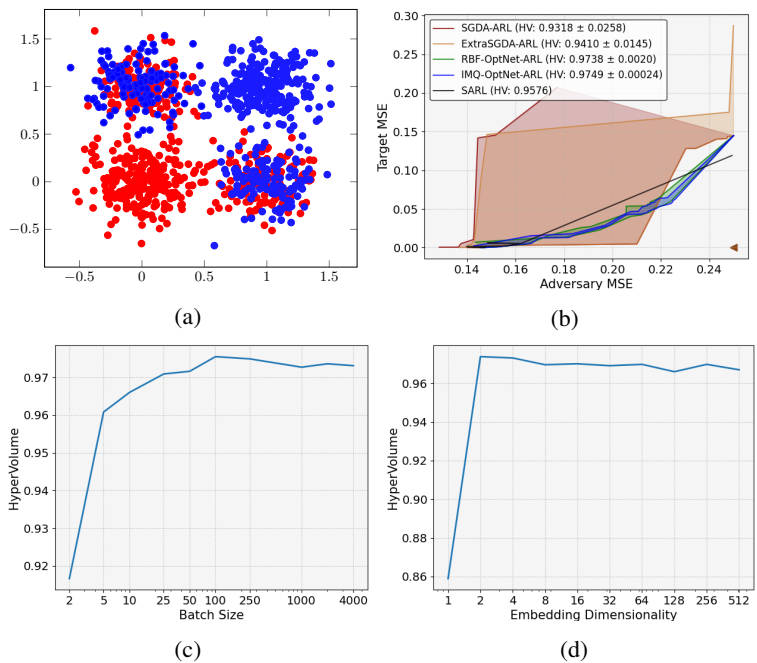
\begin{figure}[t]
\begin{subfigure}{0.41\textwidth}
  \centering
 \begin{tikzpicture}[scale=0.6]
\centering
\hspace{-1mm}
\begin{axis}
[]
        \addplot[red!100,only marks,thick, fill=red,mark=*] table {s1.txt};
        \addplot[blue!90,only marks,thick, fill=blue,mark=*] table {s2.txt};
        \addplot[red!100, only marks, thick, fill=red, mark=*] table {y1.txt};
        \addplot[blue!90, only marks, thick, mark=*] table {y2.txt};
\end{axis}
\end{tikzpicture}\label{fig-gaussian-data}
\caption{}
\end{subfigure}
\hspace{-2mm}
\begin{subfigure}{0.41\textwidth}
\includegraphics[width=\textwidth]{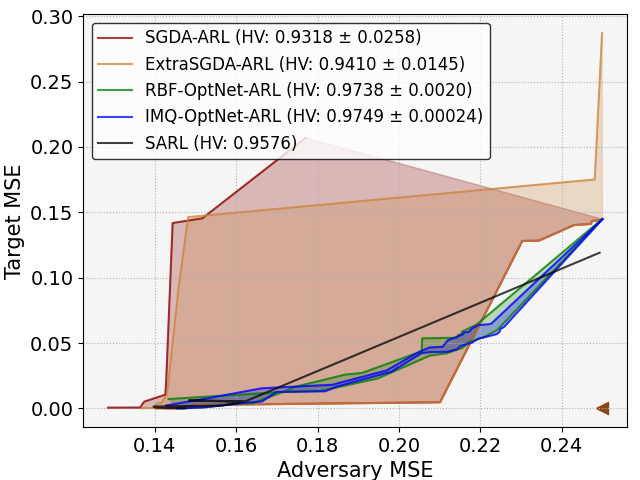}
\caption{}
\end{subfigure}

\begin{subfigure}{0.41\textwidth}
\includegraphics[width=\textwidth]{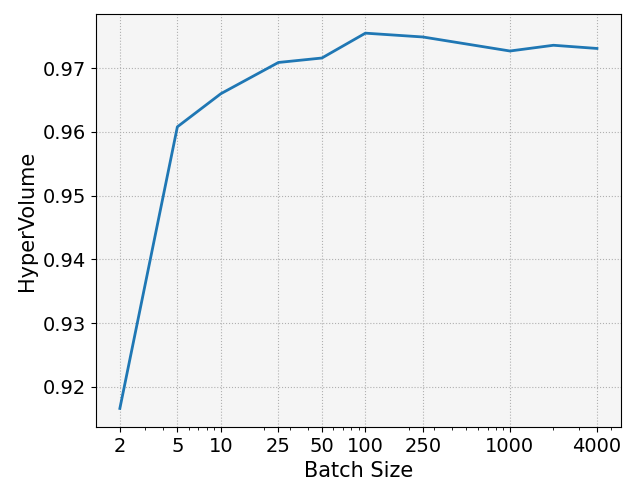}
\caption{}
\end{subfigure}
\begin{subfigure}{0.41\textwidth}
\includegraphics[width=\textwidth]{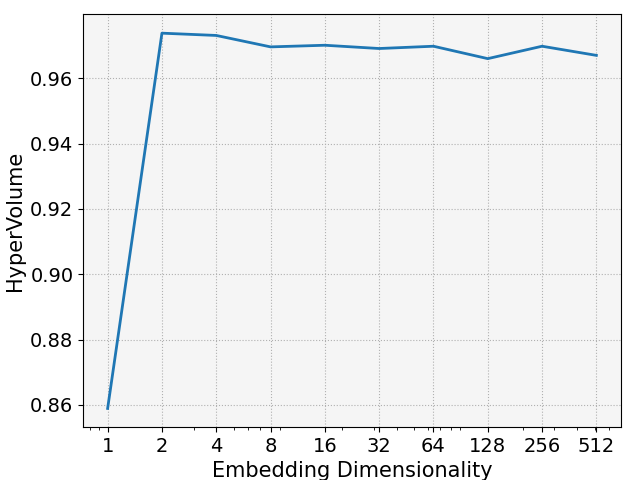}
\caption{}
\end{subfigure}
\caption{\textbf{Mixture of Gaussians:} (a) Input data. The target task is to learn a representation which is informative enough to reconstruct the input data and at the same time hide the color information ({\color{blue}$\bullet$} vs {\color{red}$\bullet$}). (b) Trade-off between the MSEs of adversary and target task for different ARL methods. (c) The HVs of OptNet-ARL (Gaussian kernel) vs different batch size values in $[2, 4000]$. (d) The HV values of OptNet-ARL (Gaussian kernel) vs different values of $r$ in $[1, 512]$. 
}\label{fig-results}
\end{figure}

In this experiment we consider a simple example where the data is generated by a mixture of four different Gaussian distributions. Let $\{f_i\}_{i=1}^4$ be all Gaussian distributions with means at $(0, 0)$, $(0, 1)$, $(1, 0)$, and $(1, 1)$, respectively and covariance matrices all equal to $\bm \Sigma = 0.2^2 \bI_2$.
Denote by $f(\x)$ the distribution of input data. Then, 
\begin{eqnarray}
&&f(\x|\,{\color{red}\bullet}) = f_1(\x) + \frac{1}{2} f_2(\x) + + \frac{1}{2} f_3(\x),\quad  \quad P\{{\color{red}\bullet}\}=\frac{1}{2}\nn \\
&&f(\x|\,{\color{blue}\bullet}) = f_4(\x) + \frac{1}{2} f_2(\x) + + \frac{1}{2} f_3(\x),\quad  \quad P\{{\color{blue}\bullet}\}=\frac{1}{2}\nn
\end{eqnarray}
 The sensitive attribute is assumed to be the color ($0$ for red and $1$ for blue) and the target task is reconstructing the input data.
 We sample $4000$ points for training and $1000$ points for testing set independently. For visualization, the testing set is shown in Figure~\ref{fig-results}(a). In this illustrative dataset , the correlation between input data and color is $0.61$ and therefore there is no encoder which results in full target performance at no leakage of sensitive attribute. Network architecture details are shown in Table~\ref{tbl-settings}.

\noindent {\bf{Stability and Trade-off}:} Figure~\ref{fig-results}(b) illustrates the five-run attainment surfaces and median hypervolumes for all methods. Since the the dimensionality of both input and output is $2$, the optimal embedding dimensionality is equal to $2$ which we set it in this experiment. 
%The solid curves demonstrate the median of five run for each method and the shadowed area around the curves represent the variance.
We note that SARL achieves hypervolume better than SGDA and ExtraSGDA ARLs which is not surprising due to the strong performance of SARL on small-sized datasets. 
However, SARL is not applicable to large datasets.
% We observe that both the iterative baselines SGDA-ARL and ExtraSGDA-ARL have high variance in the final objectives, demonstrating the instability and unreliability of such approaches. 
 Among other baselines, ExtraSGDA-ARL appears to be slightly better. In contrast, the solutions obtained by RBF-OptNet-ARL (Gaussian kernel) outperform all baselines and are highly stable across different runs, which can be observed from both attainment surfaces and hypervolumes. Addition to Gaussian kernel, we also used inverse multi quadratic (IMQ) kernel~\cite{souza2010kernel}\footnote{$k(\bm z, \bm z^\prime)=\frac{1}{\sqrt{\|\bm z-\bm z^\prime\|^2+c^2}}$} for OptNet to examine the effect kernel of function. As we observe from Figure~\ref{fig-results}(b), IMQ-OptNet-ARL performs almost similar to 
 OptNet-ARR with Gaussian kernel in terms of both trade-off and stability. 

\noindent {\textbf{Batch Size:}}
In order to examine the effect of batch size on OptNet-ARL (with Gaussian kernel), we train the encoder with different values of batch size between $2$ and $4000$ (entire training data). The results are illustrated in Figure~\ref{fig-results}(c). We observer that the trade-off HV is quite insensitive to batch sizes greater than $25$ which implies that the gradient of min-batch is an accurately enough estimator of the gradient of entire data.

\noindent {\textbf{Embedding Dimensionality:}}
We also study the effect of embedding dimensionality ($r$) by examining
different values for $r$ in $[1, 512]$ using RBF-OptNet-ARL. The results are illustrated in Figure~\ref{fig-results}(d). It is evident that the optimal embedding dimensionality ($r=2$) outperforms other values of $r$. Additionally, HV of $r=1$ suffers severely due to the information loss in embedding, while for $2<r\le 512$ the trade-off performance is comparable to that of optimal embedding dimensionality, $r=2$.  

\section{Concluding Remarks}

Adversarial representation learning is a minimax theoretic game formulation that affords explicit control over unwanted information in learned data representations. Optimization algorithms for ARL such as stochastic gradient descent-ascent (SGDA) and their variants are sub-optimal, unstable and unreliable in practice. In this paper, we introduced OptNet-ARL to address this challenge by employing differentiable closed-form solvers, such as kernelized ridge regressors, to model the ARL players that are downstream from the representation. OptNet-ARL reduces iterative SGDA to a simple optimization, leading to a fast, stable and reliable algorithm that out-performs existing ARL approaches on both small and large scale datasets. 

\vspace{5pt}
\noindent\textbf{Acknowledgements:} This work was performed under the following financial assistance award 60NANB18D210 from U.S. Department of Commerce, National Institute of Standards and Technology.

\bibliographystyle{ieeetr}

% \small{
% % \nocite{*}
% \bibliographystyle{plain}
% \bibliography{mybib}

\begin{thebibliography}{8}
\bibitem{agrawal2019differentiable}
Agrawal, A., Amos, B., Barratt, S., Boyd, S., Diamond, S.,  Kolter, Z. Differentiable convex optimization layers. In: Advances in Neural Information Processing Systems (2019)
\bibitem{amos2017optnet}
Amos, B., Kolter, J. Z. Optnet: Differentiable optimization as a layer in neural networks. In: International Conference on Machine Learning (2017)
\bibitem{balduzzi2018mechanics}
Balduzzi, D., Racaniere, S., Martens, J., Foerster, J., Tuyls, K., Graepel, T. The mechanics of n-player differentiable games. In: International Conference on Machine Learning (2018)
\bibitem{bertinetto2018meta}
Bertinetto, L., Henriques, J. F., Torr, P. H. Meta-learning with differentiable closed-form solvers. In: International Conference on Learning Representations (2018)
\bibitem{bertran2019adversarially}
Bertran, M., Martinez, N., Papadaki, A., Qiu, Q., Rodrigues, M., Reeves, G.,  Sapiro, G. Adversarially learned Representations for information obfuscation and inference. In: International Conference on Machine Learning (2019)
\bibitem{beutel2017data}
Beutel, A., Chen, J., Zhao, Z., Chi, E. H. Data decisions and theoretical implications when adversarially learning fair representations. In: Fairness,
Accountability, and Transparency in Machine Learning (2017)
\bibitem{creager2019flexibly}
Creager, E., Madras, D., Jacobsen, J. H., Weis, M., Swersky, K., Pitassi, T.,  Zemel, R. Flexibly fair representation learning by disentanglement. In: International Conference
on Machine Learning, pp. 1436-1445 (2019)
\bibitem{daskalakis2018limit}
Daskalakis, C., Panageas, I. The limit points of (optimistic) gradient descent in min-max optimization. In: Advances in Neural Information Processing Systems (2018) 
\bibitem{dua2017uci}
UCI machine learning repository, \url{http://archive.ics.uci.edu/ml}.
\bibitem{edwards2015censoring}
Edwards, H.,  Storkey, A. Censoring representations with an adversary. In: International Conference on Learning Representations (2015)
\bibitem{elazar2018adversarial}
Elazar, Y.,  Goldberg, Y. Adversarial removal of demographic attributes from text data. In: Empirical Methods in Natural Language Processing (2018)
\bibitem{ganin2015unsupervised}
Ganin, Y.,  Lempitsky, V. Unsupervised domain adaptation by backpropagation. In: International Conference on Machine Learning, pp. 1180-1189 (2015)
\bibitem{ganin2016domain}
Ganin, Y., Ustinova, E., Ajakan, H., Germain, P., Larochelle, H., Laviolette, F.,Marchand, M., Lempitsky, V. Domain-adversarial training of neural networks. In: Journal of Machine Learning Research, \textbf{17} (1), pp. 2096-2030 (2016)
\bibitem{gidel2018variational}
Gidel, G., Berard, H., Vignoud, G., Vincent, P.,  Lacoste-Julien, S. A variational inequality perspective on generative adversarial networks. In: International Conference on Learning Representations (2019)
\bibitem{golub1973differentiation}
Golub, G. H., Pereyra, V. The differentiation of pseudo-inverses and nonlinear least squares problems whose variables separate. In: SIAM Journal on Numerical Analysis, \textbf{10} (2), pp 413-432 (1973)
\bibitem{gretton2005kernel}
Gretton, A., Herbrich, R., Smola, A., Bousquet, O.,  Sch{\"o}lkopf, B. Kernel methods for measuring. In: Journal of Machine Learning Research independence, \textbf{6} , pp. 2075-2129 (2005)
\bibitem{ionescu2015training}
Ionescu, C., Vantzos, O.,  Sminchisescu, C. Training deep networks with structured layers by matrix backpropagation. In: IEEE International Conference on Computer Vision (2015)
\bibitem{jacot2018neural}
Jacot, A., Gabriel, F.,  Hongler, C. Neural tangent kernel: Convergence and generalization in neural networks. In: Advances in Neural Information Processing Systems (2018)
\bibitem{jin2019local}
Jin, C., Netrapalli, P., Jordan, M. What is local optimality in nonconvex-nonconcave minimax Optimization ? arXiv preprint arXiv:1902.00618 (2019)
\bibitem{kim2019learning}
Kim, B., Kim, H., Kim, K., Kim, S.,  Kim, J. Learning not to learn: Training deep neural networks with biased data. In: IEEE Conference on Computer Vision and Pattern Recognition (2019)
\bibitem{kingma2013auto}
Kingma, D. P.,  Welling, M. Auto-encoding variational bayes. arXiv preprint arXiv:1312.6114 (2013)
\bibitem{knowles2005summary}
Knowles, J. A summary-attainment-surface plotting method for visualizing the performance of stochastic multiobjective optimizers. In: International Conference on Intelligent Systems Design and Applications (2015)
\bibitem{korpelevich1976extragradient}
Korpelevich, G. M. The extragradient method for finding saddle points and other problems. In: Matecon, \textbf{12}, pp. 747-756 (1976)
\bibitem{kumar2012sampling}
Kumar, S., Mohri, M.,  Talwalkar, A. Sampling methods for the {N}ystr{\"o}m Method. In: Journal of Machine Learning Research, \textbf{13} (4), pp.981-1006 (2012)
\bibitem{lee2019meta}
Lee, K., Maji, S., Ravichandran, A.,  Soatto, S. Meta-learning with differentiable convex optimization. In: IEEE Conference on Computer Vision and Pattern Recognition (2019)
\bibitem{letcher2019differentiable}
Letcher, A., Balduzzi, D., Racaniere, S., Martens, J., Foerster, J., Tuyls, K.,  Graepel, T. Differentiable game mechanics. In: Journal of Machine Learning Research, \textbf{20} (84), pp. 1-40 (2019)
\bibitem{liu2015deep}
Liu, Z., Luo, P., Wang, X.,  Tang, X. Deep learning face attributes in the wild. In: IEEE International Conference on Computer Vision (2015)
\bibitem{louizos2015variational}
Louizos, C., Swersky, K., Li, Y., Welling, M.,  Zemel, R. The variational fair autoencoder. In: arXiv preprint arXiv:1511.00830 (2015)
\bibitem{madras2018learning}
Madras, D., Creager, E., Pitassi, T.,  Zemel, R. Learning adversarially fair and transferable representations. In: International Conference on Machine Learning (2018)
\bibitem{mescheder2017numerics}
Mescheder, L., Nowozin, S.,  Geiger, A. The numerics of gans. In: Advances in Neural Information Processing Systems (2017)
\bibitem{moyer2018invariant}
Moyer, D., Gao, S., Brekelmans, R., Steeg, G. V.,  Galstyan, A. Invariant representations without adversarial training, In: Advances in Neural Information Processing Systems (2018)
\bibitem{nagarajan2017gradient}
Nagarajan, V.,  Kolter, J. Z. Gradient descent GAN optimization is locally stable. In: Advances in Neural Information Processing Systems (2017)
\bibitem{roy2019mitigating}
Roy, P. C.,  Boddeti, V. N. Mitigating information leakage in image representations: A maximum entropy approach. In: IEEE Conference on Computer Vision and Pattern Recognition (2019)
\bibitem{sadeghi2020imparting}
Sadeghi, B.,  Boddeti, V. N. Imparting fairness to pre-trained biased representations. In: IEEE Conference on Computer Vision and Pattern Recognition Workshops (2020)
\bibitem{sadeghi2019global}
Sadeghi, B., Yu, R.,  Boddeti, V. On the global optima of kernelized adversarial representation learning. In: IEEE International Conference on Computer Vision (2019)
\bibitem{shawe2004kernel}
Shawe-Taylor, J.,  Cristianini, N. Kernel methods for pattern analysis. In: Cambridge University Press (2014)
\bibitem{song2018learning}
Song, J., Kalluri, P., Grover, A., Zhao, S.,  Ermon, S.  Learning controllable fair representations. In: International Conference on Artificial Intelligence and Statistics (2019)
\bibitem{tzeng2017adversarial}
Tzeng, E., Hoffman, J., Saenko, K.,  Darrell, T. Adversarial discriminative domain adaptation. In: IEEE Conference on Computer Vision and Pattern Recognition (2017)
\bibitem{valmadre2017end}
Valmadre, J., Bertinetto, L., Henriques, J., Vedaldi, A.,  Torr, P. H. End-to-end representation learning for correlation filter based tracking. In: IEEE conference on computer vision and pattern recognition (2017)
\bibitem{xie2017controllable}
Xie, Q., Dai, Z., Du, Y., Hovy, E.,  Neubig, G.  Controllable invariance through adversarial feature learning. In: Advances in Neural Information Processing Systems (2017)
\bibitem{zemel2013learning}
Zemel, R., Wu, Y., Swersky, K., Pitassi, T.,  Dwork, C.  Learning fair representations. In: International Conference on Machine Learning (2013)
\bibitem{zhang2018mitigating}
Zhang, B. H., Lemoine, B.,  Mitchell, M. Mitigating unwanted biases with adversarial learning. In: AAAI/ACM Conference on AI, Ethics, and Society (2018)
\bibitem{zitzler1998multiobjective}
Zitzler, E.,  Thiele, L. Multiobjective optimization using evolutionary algorithms—a comparative case study. In: International conference on parallel problem solving from nature (1998)
\bibitem{fisher1936use}
Fisher, Ronald A. The use of multiple measurements in taxonomic problems. In: Annals of human eugenics. In: Wiley Online Library (1926)
\bibitem{hardt2016train}
Hardt, M.,  Recht, B., Singer, Y. Train faster, generalize better: Stability of stochastic gradient descent. In: International conference on machine learning (2016)
\bibitem{souza2010kernel}
Souza, C{\'e}sar R. Kernel functions for machine learning applications. In: Creative commons attribution-noncommercial-share alike (2016)
\end{thebibliography}
% }
\end{document}